\documentclass[runningheads]{llncs}
\usepackage{draftwatermark}
\SetWatermarkText{Preprint}
\SetWatermarkScale{4}
\SetWatermarkColor[gray]{0.90}
\usepackage{graphicx}
\usepackage{algorithmic}
\usepackage{algorithm}

\makeatletter
\newcommand{\HEADER}[1]{\ALC@it\underline{\textsc{#1}}\begin{ALC@g}}
\newcommand{\ENDHEADER}{\end{ALC@g}}
\makeatother

\usepackage{amsmath,amsfonts}

\begin{document}
\title{A General Recipe for Automated Machine Learning in Practice}
%
%
\author{Hernan Ceferino Vazquez}
\authorrunning{H. C. Vazquez}
%
\institute{MercadoLibre Inc.\\
Applied Machine Learning Research, Argentina\\
\email{hernan.vazquez@mercadolibre.com},
\email{hernan.c.vazquez@gmail.com}}
\maketitle              
\begin{abstract}
Automated Machine Learning (AutoML) is an area of research that focuses on developing methods to generate machine learning models automatically. The idea of being able to build machine learning models with very little human intervention represents a great opportunity for the practice of applied machine learning. However, there is very little information on how to design an AutoML system in practice. Most of the research focuses on the problems facing optimization algorithms and leaves out the details of how that would be done in practice. In this paper, we propose a frame of reference for building general AutoML systems. Through a narrative review of the main approaches in the area, our main idea is to distill the fundamental concepts in order to support them in a single design. Finally, we discuss some open problems related to the application of AutoML for future research.

\end{abstract}
\section{Introduction}

In last decade, machine learning has been applied in various fields and used to solve many challenging business problems. This has led to a growing demand for data scientists with solid knowledge and experience to harness massive amounts of data and create business-impacting machine learning solutions \cite{bouneffouf2020}. However apply machine learning to business problems is labor-intensive and human experts are scarce and heavily demanded in organizations.

Automated machine learning (AutoML) has become an area of growing interest for machine learning researchers and practitioners. AutoML groups together many techniques and methods that can be used to automate the tasks that constitute the process of applying machine learning. This has led the researchers to propose many literature reviews that try to to summarize the area from different perspectives and propose many reusable components to solve the different AutoML challenges.

In this work we take advantage of these perspectives to propose a general recipe that brings them closer to the practice of applied machine learning. Based on findings in several AutoML reviews, we describe a design based on learning loops that attempts to provide the flexibility to incorporate main AutoML methods. Furthermore, we propose a new way to approach the goal of AutoML systems from a multi-objective perspective that takes into account time and computational resources.

\section{Methodology}

\begin{figure}
\centering
    \includegraphics[width=350pt]{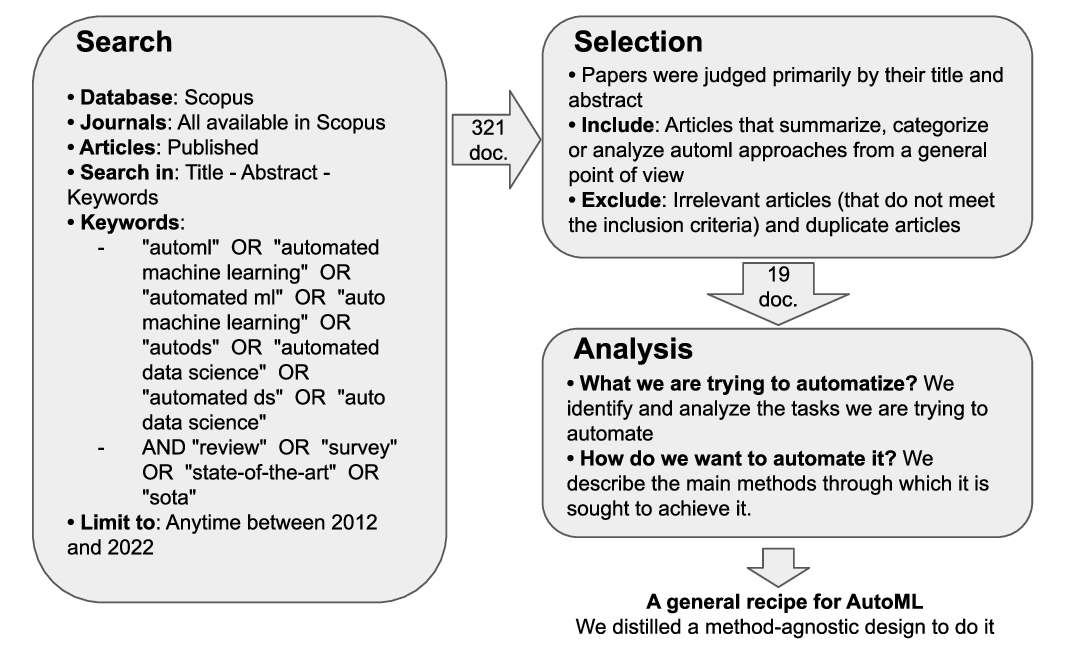}
\caption{Process flow for AutoML literature review.} \label{figure0}
\end{figure}

The paper aims to gather information on AutoML, especially when it comes from literature reviews. This is because the main objective is to identify the most important parts for the comprehensive application of AutoML systems in practice. To carry out the literature review, the narrative and scoping literature review approaches have been adopted \cite{pare2015synthesizing,cronin2008undertaking}, and a research search strategy has been developed \cite{creswell2017research,robson2002real}. Figure \ref{figure0} shows the process flow for the systematic literature review.

The articles on AutoML were identified from the Scopus database to find the most relevant published articles or in press articles. We search within the title, abstract and key words for various terms such as ”automated machine learning”, ”automl”, ”automated data science” and ”autods”. The search is then narrowed to documents that also contain either in the title or the abstract or in the keywords, the terms "review", "survey", "state-of-the-art" or "sota". With this we seek to keep all the articles that summarize various methods in the area. In order to focus on recent literature, the search is limited to articles published in the last decade. The search was carried out on May 11, 2022 and retrieved 321 documents. After manual screening, e.g. removing duplicate or irrelevant articles, 19 articles remained, which form the core of this review.

\section{Results Analysis}

In this section the results are presented. We focus on answering two main questions. The first question is: What we are trying to automatize? With this question we are trying to identify and analyze the tasks we are trying to automate. The second question is: How do we want to automate it? With this question we are looking for describing the main methods through which it is sought to achieve the automation.

\subsection{What we are trying to automatize?}

Most of the articles describe AutoML as a process in which tasks that would normally be performed by a data scientist are automated. Table \ref{tab1} describes an overall process and maps each part of the process to the articles. It is evident that the majority of articles are concentrated around \textit{Model Selection} and \textit{Hyperparameter Optimization}. Moreover, the least explored areas are \textit{Task Formulation} and \textit{Prediction Engineering}.

\begin{table}
\caption{Machine learning process phases identified in the reviewed articles.}\label{tab1}
\centering
\begin{tabular}{|l|l|}
\hline
Phases & Reviews \\
\hline
Task Formulation & \cite{santu2022,de2021automating}\\
Prediction Engineering & \cite{santu2022,de2021automating}\\
Data Preparation & \cite{he2021,lakshmi2021,escalante2021,zoller2021,kulbach2020,bouneffouf2020,de2021automating,nagarajah2019,nagarajah2019b}\\
Feature Engineering & \cite{santu2022,li2021,li2021b,he2021,lakshmi2021,escalante2021,zoller2021,kulbach2020,bouneffouf2020,waring2020,de2021automating,tuggener2019,nagarajah2019,nagarajah2019b}\\
Model Selection & \cite{santu2022,li2021,li2021b,he2021,lakshmi2021,escalante2021,zoller2021,vaccaro2021,vaccaro2020,kulbach2020,bouneffouf2020,waring2020,elshawi2020,de2021automating,tuggener2019,weng2019,nagarajah2019,nagarajah2019b}\\
Hyperparameter Opt. & \cite{santu2022,li2021,li2021b,he2021,lakshmi2021,escalante2021,zoller2021,vaccaro2021,vaccaro2020,kulbach2020,bouneffouf2020,waring2020,elshawi2020,de2021automating,tuggener2019,weng2019,nagarajah2019,nagarajah2019b}\\
Model Estimation & \cite{he2021,escalante2021,zoller2021,kulbach2020,waring2020,elshawi2020,tuggener2019}\\
Results Summarizing & \cite{santu2022,bouneffouf2020,de2021automating}\\
\hline
\end{tabular}
\end{table}


\subsubsection{Task Formulation}
This is the process through which a machine learning task is formulated that could help solve a business problem. Only two works of those reviewed incorporate this phase within the scope of AutoML. Santu et al. \cite{santu2022}, highlights in this phase the interaction between domain experts and data scientists, while De Bie et al. \cite{de2021automating} relates it more to an EDA (Exploratory Data Analysis) process. Generally, the output of this task are available data sources, verified hypotheses and the main business metric to impact.

\subsubsection{Prediction Engineering}
This is the phase in which the business problem is framed as a machine learning problem. This includes deciding between different frameworks, for example, a ranking problem can be solved as a scoring problem (point-wise), a binary classification problem (pair-wise) or a position assignment problem (list-wise). According to Santu et al. \cite{santu2022} This phase also involves constructing and assigning labels to data points according to the goal prediction task. The output of this phase is generally the framing of the problem represented by the data points and targets, and a refinement of the business metric into a proxy metric to be optimized.

\subsubsection{Data Preparation}
Many jobs incorporate data preparation as part of the tasks that can be automated. The data preparation process consists of performing operations on the defined dataset to make it ready for the next process. Within this process we define two types of data preparation, those that increase the number of data points (e.g. data collection, data augmentation) \cite{he2021,lakshmi2021,escalante2021,bouneffouf2020,de2021automating,nagarajah2019} and those that do not (e.g. data cleaning, data inputation, data standardization) \cite{he2021,lakshmi2021,zoller2021,kulbach2020,bouneffouf2020,de2021automating,nagarajah2019,nagarajah2019b}. The output of this phase is typically a curated dataset ready to be used for feature engineering.

\subsubsection{Feature Engineering}

Feature Engineering task aims to maximize the extraction of
features from raw data for use by algorithms and models\cite{he2021}. In this context, raw data could be structured data, such as tabular and relational datasets \cite{bouneffouf2020}, or unstructured data, such as text and images. Feature Engineering consists mainly of two sub-task: feature
selection and feature transformation (e.g. feature extraction, feature construction). Feature Engineering is one of the most explored task due to its impact on the performance of the model \cite{zoller2021} since data and features determine the upper bound of ML, and that models and algorithms can only approximate this limit.

\subsubsection{Model Exploration and Hyper-parameter Tuning}
This task is one of the most explored in the literature as it was one of the places where researchers started looking for automation \cite{feurer2019hyperparameter}. With the advancement in computing power, a growing wave of machine learning methods and techniques became available to data scientists. Being able to explore different models with different hyper-parameters automatically is something that usually saves machine learning practitioners a lot of time. This problem is generally approached from two sub-problems: The definition of a search space of possible models to be explored and the search method to be used to traverse that space.

\subsubsection{Model Estimation}
Evaluating models is an expensive process, since it usually requires a series of training and test stages, usually in a cross-validation scenario using all the data. Because of this, researchers have focused over the years on creating methods to estimate the performance of models in less expensive ways \cite{he2021}. This is usually done in roughly two ways. Either reducing the amount of data needed to evaluate the model (e.g. multi-fidelity approaches \cite{fernandez2016review}) or modeling the performance of the models in a way that allows us to predict it without the need to evaluate it (e.g. surrogate models \cite{eggensperger2014surrogate}, relative landmarks \cite{fusi2018probabilistic}). 

\subsubsection{Results Summarizing/Recommendation}
The last part of the procedures is to summarize all the findings and recommend the most useful/promising solution to the stakeholders. There is very little information about this task in the literature. Santu et al \cite{santu2022} consider that the recommendations are made at the model, function or computational overhead level. This part is still mostly done manually without any systematic structure. However, some AutoML tools automatically select the best solution from the target metric, while others allow the data scientist to select an option from a ranking of available options.

\subsection{How do we want to automate it?}

The general way researchers have found to automate the process is to think of it as a search problem. Every possible decision within the machine learning application process becomes a configuration variable. Thus, the problem is reduced to finding the best configuration among all possible configurations. The main methods to carry out this search according to the review of the literature are described below.

\subsubsection{Random Search and Grid Search}
Random Search and Grid Search are the most widely used strategies for automatically explore the search space for hyper-parameter optimization \cite{bergstra2012random}.Random search consists of exploring the search space randomly. Usually this search is restricted to a fixed number of attempts. Grid Search, consists of exploring the search space as if it were a grid. For this it is necessary to discretize the values of the continuous numerical variables in order to fit them into the grid. Both methods are widely used but do not have any kind of optimization when it comes to exploring the space efficiently.

\subsubsection{SMBO/SMAC}
Sequential Model-Based Optimization (SMBO) involves tuning a model of the predictive performance at the same time as configurations are explored \cite{jones1998efficient}. It then uses it to make decisions about which configurations are most promising to evaluate. The classical implementation of this model is using Bayesian optimization and a surrogate model based on Gaussian processes. In Sequential Model-based Algorithm Configuration (SMAC) \cite{hutter2011sequential} Hutter et al generalize this model in an attempt to overcome some of its limitations. They do this mainly by using random forests as surrogate models.

\subsubsection{Reinforcement Learning}
Reinforcement Learning has been widely used as a search method \cite{he2021}. It mainly consists of a controller model, usually a recurrent neural network (RNN) \cite{zoph2018learning}. The controller executes an action at each step to sample a new configuration from the search space and receives an observation of the state together with a reward from the environment to update the controller’s sampling strategy. Here environment refers to the application of the configuration to the training procedure to train and evaluate the solution generated by the controller, after which the corresponding predictive performance (such as accuracy) are returned.

\subsubsection{Evolution Based Methods}
Evolution-based optimization methods follow a process inspired by biological concepts related to evolution \cite{nagarajah2019b}. Generally the most used is the one based on genetic programming. In this method, it first creates a random population of possible configurations from the search space. Then each individual (configuration) of the population is evaluated to know its fitness function (predictive performance). Based on this aptitude, the best builds have a higher chance of passing through to the next generation and interbreeding with others. Generally this process is repeated until the performance is not improved or until a certain number of generations is reached.

\subsubsection{Bandit-based methods}
Bandit-based methods consist of dividing the search space and evaluating many options in parallel to then decide how to proceed \cite{elshawi2020}. The two most popular strategies are Successive Halving \cite{jamieson2016non} and HyperBand \cite{li2017hyperband}. On the one hand, Successive Halving consists of first evaluating all configurations with a small data set. Configurations are then ranked based on their performance and the worst half is eliminated. Finally, the data is doubled and the process is repeated until only one configuration remains. On the other hand, HyperBand uses the same technique, but instead of eliminating less promising configurations, it assigns them a lower chance of being selected in the next iteration.

\subsubsection{Adaptive methods}
Adaptive methods are those that aim to adapt the configuration during training. This type of method is commonly used in neural architecture search (NAS) to learn the best network architecture while learning its parameters \cite{he2021}. For example, self-tuning networks (STN) and population based training (PBT) fall into this category. Furthermore, in deep learning, another widely used method is to adapt the learning rate during the training of a network \cite{zeiler2012adadelta}.

\subsubsection{Meta-learning}
Meta-learning is learned from prior experience in a systematic, data-driven way \cite{vanschoren2019meta}. It is a process which can be found in many reviews about AutoML and that aims to improve the process itself from learning obtained after the application in many tasks. It generally consists of two problems. The first is how to represent and collect the prior knowledge, usually through meta-features. The second problem is how to learn from this data to extract and transfer knowledge that guides the process of finding an optimal solution for new tasks. Meta-learning techniques can generally be roughly categorized into three broad groups \cite{elshawi2020}: learning based on the properties of the task, learning from evaluations of previous models, and learning from already trained models.

\section{A General Recipe for AutoML}

\begin{figure}
\includegraphics[width=\textwidth]{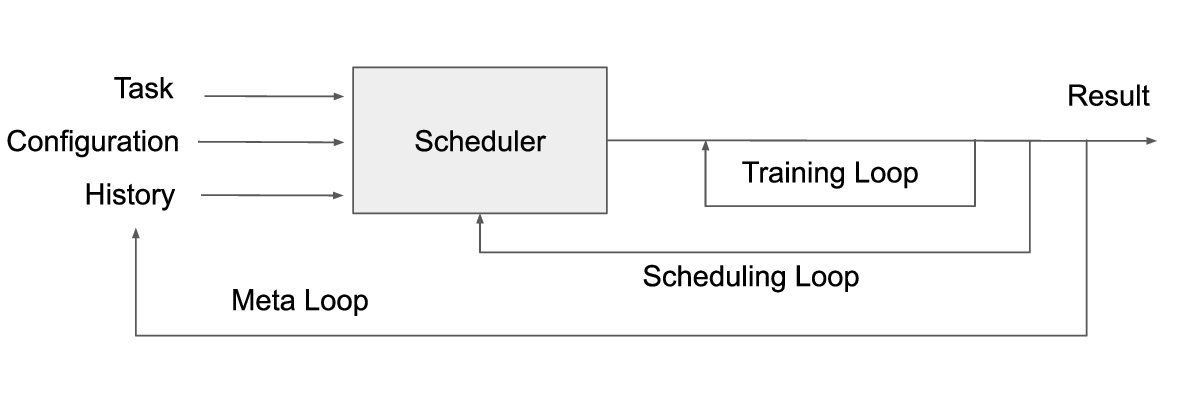}
\caption{Context diagram of the main feedback loops in AutoML.} \label{figure1}
\end{figure}

There are several AutoML methods in literature that could be used to search the best machine learning solution to an specific problem. Most of this methods rely on a feedback loop to explore efficiently the search space. In particular, we identify the necessity of three main loops in which the search of the best machine learning solution can be decomposed (Figure \ref{figure1}). Each of these learning loops are described below.

Another important component of AutoML systems is the objective function. The objective function is the function that the system aims to maximize or minimize. We will describe these components in more detail in the following sections.

\subsection{Scheduling Loop}
AutoML systems rely on the possibility of evaluating a possible good configuration, analyzing the results and being able to decide which is the next best configuration to evaluate. This is the basis for most AutoML methods. We will call the component that is responsible for making the decision the scheduler. Algorithm \ref{alg1} shows the pseudo-code for the general operation of the scheduler.
The scheduler is not only responsible for deciding which are the best configurations to explore but also for defining an evaluation plan to efficiently explore the search space. This evaluation plan is composed of two abstractions, the steps and the stages. A step can be defined as the evaluation of one configuration. A stage can be defined as a set of independent steps that are likely to be parallelized. For example, Random Search and Grid Search only contain steps. SMBO and SMAC only contain 1-step Stages. Reinforcement learning techniques are also commonly 1-Step Stages. Bandit-based and Evolution-based methods, require the evaluation of a set of stages made up of independent steps.
In real-world scenarios, the scheduler may also have to decide on which computational resources to perform the evaluation and for how long (for those methods that may not converge) \cite{feurer2020auto}.

\begin{algorithm}
  \caption{AutoML Learning Algorithm}\label{alg1}
  \begin{algorithmic}
  \STATE \textbf{Input}: Tasks, Configurations, History, Resources
  \FOR{task in Tasks}
     \STATE Create an Evaluation Plan consisting of a set of stages,
     \STATE \quad with each stage being made up of independent steps.
     \STATE \quad with each independent step assigned to a resource
      \FOR{stage in Evaluation Plan}
          \FOR{step in stage}
            \STATE Evaluate the step
            \STATE Add step results to the stage results
          \ENDFOR
      \STATE Update the Evaluation Plan with stage results
      \ENDFOR
    \STATE Summarize results     
    \STATE Add results to History 
  \ENDFOR
\end{algorithmic}
\end{algorithm}

\subsection{Meta Loop}
Another important capability of an AutoML system is to be able to learn from its own experience and thus become more and more efficient in exploration. We describe this learning ability as the Meta-Loop. The Meta-Loop allows us to learn transversely about the problems we are solving. This loop is potentially exploited by meta-learning techniques that learn from other tasks. Something important to enable this learning is to define how the information will be stored, what will be the way in which the tasks will be described (meta-features) and how this information will be consumed by the scheduler. Many authors have studied how to represent tasks for use in a machine learning process. On the one hand, some works have made an effort to identify the best characteristics that describe a data set \cite{rivolli2018characterizing}. On the other hand, others have chosen to create distributed representations \cite{achille2019task2vec,jomaa2021dataset2vec}.

\subsection{Training Loop}
Finally, it is possible see a third loop, the training loop. This loop occurs in training time, and it is the basis for adaptive methods that change configurations in a single trial, like adaptive learning rate \cite{zeiler2012adadelta}. In adaptive learning rate, the value is selected in a dynamic way using information in training time. This attempts to alleviate the task of choosing the best learning rate before training.
In practice, the learning information may not return to the scheduler until the training is complete. This is due to the overhead that can be caused if the scheduler and the training are running on different processes or even machines. Because of this, it is very important to consider that the scheduler will only see, for example, that adaptive learning rate was activated (as a binary configuration) and then see the results of this after training.

\subsection{Objective Function}
The AutoML problem is generally defined as a combined algorithm selection and hyperparameter search (CASH) problem \cite{kotthoff2019auto}. In CASH the objective is to find a pipeline ($\mathcal{M}$) and a set of hyperparameters ($\lambda$) that minimizes the generalization error ($GE$) of a particular task ($\mathcal{D}$). Feurer et al. extend this definition to generalize it to many tasks and thus include the idea of meta-learning in the optimization problem \cite{feurer2020auto}. It also proposes incorporating time and computational resources as constraints on how much we are willing to invest ($T$) as shown in Equation 1.

\begin{equation}\label{eq1}
\mathcal{M}_{{\lambda}^*}\in \underset{\lambda\in\Lambda}{\text{argmin}}\;\widehat{GE}(\mathcal{M}_\lambda,\mathcal{D}) \quad \text{s.t.} \quad (\sum_{}^{} {t_\lambda}_i) < T
\end{equation}

This definition is very useful at the experimental level. However, in practice, two solutions can achieve the same or similar predictive performance and consume far fewer resources. That solution would probably be the best. In that case, seeing the budget as a constraint is not useful. Based on the works reviewed, we believe that it is most convenient to model the problem as a multi-objective optimization problem, in which the aim is to minimize the generalization error together with the time and computational resources used.

\begin{equation}
\mathcal{M}_{{\lambda}^*}\in \underset{\lambda\in\Lambda}{\text{argmin}}\;(\widehat{GE}(\mathcal{M}_\lambda,\mathcal{D} )
\wedge (\sum_{}^{} {t_\lambda}_i))
\quad \text{s.t.} \quad (\sum_{}^{} {t_\lambda}_i) < T
\end{equation}

where $\mathcal{M}_{{\lambda}^*}$ denotes the best pipeline configuration, and ${t_\lambda}_i$ denotes the time and computational resources used to evaluate the i-th configuration $\lambda_i$ of a particular pipeline.

Equation 2 tries to synthesize the purpose of the three learning cycles presented in the previous sections. In essence, what we pursue in the general learning process is to be more and more efficient in the search for the best machine learning solutions.

\section{Discussion}

AutoML is an area that has gained importance in recent years and has led to the appearance of numerous literature reviews. In particular, we took into account only those whose sources are indexed by Scopus, leaving out gray literature that could be important. We believe that this helped us to better define the scope of this work and to define a clear methodology. In addition, we indirectly consider the references of the analyzed works where some references to gray literature were found.

The loops described in this work are important to visualize where learning occurs. On one hand, we believe that the boundaries between these loops are permeable in terms of hyper-parameters. For example, one hyper-parameter could be initialized as a range from knowledge in the meta loop and then refined in the other loops. On the other hand, this boundaries are clearly defined in terms of the execution of the loops. For example, each iteration of a higher level loop might depend on the set of iterations of lower level loops to complete.

Another interesting point to discuss is that when we consider the entire data science process for the application of machine learning, \textit{Task Formulation} and \textit{Prediction Engineering} are two of the most difficult data science task to automate. However, we believe that this general view of AutoML could support some parts of these tasks as long as the decisions made can be coded as configurations.

In addition, this article proposes to address CASH problems from a multi-objective perspective. This brings with it the new challenge of having to define the trade-off between maximizing predictive performance or minimizing time and resource consumption. This may be difficult to determine in practice and further research is required to define suitable criteria.

\section{Conclusions}

In this paper, we propose a general recipe for AutoML systems in practice generated from the findings of a systematic literature review. In particular, we describe the main tasks in the process of apply machine learning and the main methods used to automate it. After the review, we describe a general design for AutoML systems from the perspective of feedback loops necessary for learning. Additionally, we propose a multi-objective function as the general purpose for AutoML systems in practice that takes into account time and computational resources. Despite the recentness of the AutoML area, we hope this work would be helpful for research scholars and practitioners of machine learning, to understand and integrate the latest research efforts related to AutoML into your own systems.

%
%
%
%
\bibliographystyle{splncs04}
\bibliography{references}

\begin{thebibliography}{10}
\providecommand{\url}[1]{\texttt{#1}}
\providecommand{\urlprefix}{URL }
\providecommand{\doi}[1]{https://doi.org/#1}

\bibitem{achille2019task2vec}
Achille, A., Lam, M., Tewari, R., Ravichandran, A., Maji, S., Fowlkes, C.C.,
  Soatto, S., Perona, P.: Task2vec: Task embedding for meta-learning. In:
  Proceedings of the IEEE/CVF International Conference on Computer Vision. pp.
  6430--6439 (2019)

\bibitem{bergstra2012random}
Bergstra, J., Bengio, Y.: Random search for hyper-parameter optimization.
  Journal of machine learning research  \textbf{13}(2) (2012)

\bibitem{bouneffouf2020}
Bouneffouf, D., Aggarwal, C., Hoang, T., Khurana, U., Samulowitz, H., Buesser,
  B., Liu, S., Pedapati, T., Ram, P., Rawat, A., Wistuba, M., Gray, A.: Survey
  on automated end-to-end data science? In: Proceedings of the International
  Joint Conference on Neural Networks (2020), \url{www.scopus.com}, cited By :2

\bibitem{creswell2017research}
Creswell, J.W., Creswell, J.D.: Research design: Qualitative, quantitative, and
  mixed methods approaches. Sage publications (2017)

\bibitem{cronin2008undertaking}
Cronin, P., Ryan, F., Coughlan, M.: Undertaking a literature review: a
  step-by-step approach. British journal of nursing  \textbf{17}(1),  38--43
  (2008)

\bibitem{de2021automating}
De~Bie, T., De~Raedt, L., Hern{\'a}ndez-Orallo, J., Hoos, H.H., Smyth, P.,
  Williams, C.K.: Automating data science: Prospects and challenges. arXiv
  preprint arXiv:2105.05699  (2021)

\bibitem{eggensperger2014surrogate}
Eggensperger, K., Hutter, F., Hoos, H.H., Leyton-Brown, K.: Surrogate
  benchmarks for hyperparameter optimization. In: MetaSel@ ECAI. pp. 24--31
  (2014)

\bibitem{elshawi2020}
Elshawi, R., Sakr, S.: Automated Machine Learning: Techniques and Frameworks,
  Lecture Notes in Business Information Processing, vol.~390 (2020),
  \url{www.scopus.com}, cited By :1

\bibitem{escalante2021}
Escalante, H.J.: Automated Machine Learning—A Brief Review at the End of the
  Early Years. Natural Computing Series (2021), \url{www.scopus.com}

\bibitem{fernandez2016review}
Fern{\'a}ndez-Godino, M.G., Park, C., Kim, N.H., Haftka, R.T.: Review of
  multi-fidelity models. arXiv preprint arXiv:1609.07196  (2016)

\bibitem{feurer2020auto}
Feurer, M., Eggensperger, K., Falkner, S., Lindauer, M., Hutter, F.:
  Auto-sklearn 2.0: The next generation. arXiv preprint arXiv:2007.04074
  \textbf{24} (2020)

\bibitem{feurer2019hyperparameter}
Feurer, M., Hutter, F.: Hyperparameter optimization. In: Automated machine
  learning, pp. 3--33. Springer, Cham (2019)

\bibitem{fusi2018probabilistic}
Fusi, N., Sheth, R., Elibol, M.: Probabilistic matrix factorization for
  automated machine learning. Advances in neural information processing systems
   \textbf{31} (2018)

\bibitem{he2021}
He, X., Zhao, K., Chu, X.: Automl: A survey of the state-of-the-art.
  Knowledge-Based Systems  \textbf{212} (2021), \url{www.scopus.com}, cited By
  :155

\bibitem{hutter2011sequential}
Hutter, F., Hoos, H.H., Leyton-Brown, K.: Sequential model-based optimization
  for general algorithm configuration. In: International conference on learning
  and intelligent optimization. pp. 507--523. Springer (2011)

\bibitem{jamieson2016non}
Jamieson, K., Talwalkar, A.: Non-stochastic best arm identification and
  hyperparameter optimization. In: Artificial intelligence and statistics. pp.
  240--248. PMLR (2016)

\bibitem{jomaa2021dataset2vec}
Jomaa, H.S., Schmidt-Thieme, L., Grabocka, J.: Dataset2vec: Learning dataset
  meta-features. Data Mining and Knowledge Discovery  \textbf{35}(3),  964--985
  (2021)

\bibitem{jones1998efficient}
Jones, D.R., Schonlau, M., Welch, W.J.: Efficient global optimization of
  expensive black-box functions. Journal of Global optimization
  \textbf{13}(4),  455--492 (1998)

\bibitem{kotthoff2019auto}
Kotthoff, L., Thornton, C., Hoos, H.H., Hutter, F., Leyton-Brown, K.:
  Auto-weka: Automatic model selection and hyperparameter optimization in weka.
  In: Automated Machine Learning, pp. 81--95. Springer, Cham (2019)

\bibitem{kulbach2020}
Kulbach, C., Philipp, P., Thoma, S.: Personalized automated machine learning,
  Frontiers in Artificial Intelligence and Applications, vol.~325 (2020),
  \url{www.scopus.com}

\bibitem{lakshmi2021}
Lakshmi~Patibandla, R.S.M., Srinivas, V.S., Mohanty, S.N., Ranjan~Pattanaik,
  C.: Automatic machine learning: An exploratory review. In: 2021 9th
  International Conference on Reliability, Infocom Technologies and
  Optimization (Trends and Future Directions), ICRITO 2021 (2021),
  \url{www.scopus.com}

\bibitem{li2017hyperband}
Li, L., Jamieson, K., DeSalvo, G., Rostamizadeh, A., Talwalkar, A.: Hyperband:
  A novel bandit-based approach to hyperparameter optimization. The Journal of
  Machine Learning Research  \textbf{18}(1),  6765--6816 (2017)

\bibitem{li2021b}
Li, Y., Wang, Z., Ding, B., Zhang, C.: Automl: A perspective where industry
  meets academy. In: Proceedings of the ACM SIGKDD International Conference on
  Knowledge Discovery and Data Mining. pp. 4048--4049 (2021),
  \url{www.scopus.com}

\bibitem{li2021}
Li, Y., Wang, Z., Xie, Y., Ding, B., Zeng, K., Zhang, C.: Automl: From
  methodology to application. In: International Conference on Information and
  Knowledge Management, Proceedings. pp. 4853--4856 (2021),
  \url{www.scopus.com}, cited By :1

\bibitem{nagarajah2019b}
Nagarajah, T., Poravi, G.: An extensive checklist for building automl systems.
  In: CEUR Workshop Proceedings. vol.~2360 (2019), \url{www.scopus.com}

\bibitem{nagarajah2019}
Nagarajah, T., Poravi, G.: A review on automated machine learning (automl)
  systems. In: 2019 IEEE 5th International Conference for Convergence in
  Technology, I2CT 2019 (2019), \url{www.scopus.com}, cited By :10

\bibitem{pare2015synthesizing}
Par{\'e}, G., Trudel, M.C., Jaana, M., Kitsiou, S.: Synthesizing information
  systems knowledge: A typology of literature reviews. Information \&
  Management  \textbf{52}(2),  183--199 (2015)

\bibitem{rivolli2018characterizing}
Rivolli, A., Garcia, L.P., Soares, C., Vanschoren, J., de~Carvalho, A.C.:
  Characterizing classification datasets: a study of meta-features for
  meta-learning. arXiv preprint arXiv:1808.10406  (2018)

\bibitem{robson2002real}
Robson, C.: Real world research: A resource for social scientists and
  practitioner-researchers. Wiley-Blackwell (2002)

\bibitem{santu2022}
Santu, S.K.K., Hassan, M.M., Smith, M.J., Xu, L., Zhai, C., Veeramachaneni, K.:
  Automl to date and beyond: Challenges and opportunities. ACM Computing
  Surveys  \textbf{54}(8) (2022), \url{www.scopus.com}, cited By :2

\bibitem{tuggener2019}
Tuggener, L., Amirian, M., Rombach, K., Lorwald, S., Varlet, A., Westermann,
  C., Stadelmann, T.: Automated machine learning in practice: State of the art
  and recent results. In: Proceedings - 6th Swiss Conference on Data Science,
  SDS 2019. pp. 31--36 (2019), \url{www.scopus.com}, cited By :18

\bibitem{vaccaro2020}
Vaccaro, L., Sansonetti, G., Micarelli, A.: Automated Machine Learning:
  Prospects and Challenges, Lecture Notes in Computer Science (including
  subseries Lecture Notes in Artificial Intelligence and Lecture Notes in
  Bioinformatics), vol. 12252 LNCS (2020), \url{www.scopus.com}, cited By :1

\bibitem{vaccaro2021}
Vaccaro, L., Sansonetti, G., Micarelli, A.: An empirical review of automated
  machine learning. Computers  \textbf{10}(1),  1--27 (2021),
  \url{www.scopus.com}, cited By :7

\bibitem{vanschoren2019meta}
Vanschoren, J.: Meta-learning. In: Automated Machine Learning, pp. 35--61.
  Springer, Cham (2019)

\bibitem{waring2020}
Waring, J., Lindvall, C., Umeton, R.: Automated machine learning: Review of the
  state-of-the-art and opportunities for healthcare. Artificial Intelligence in
  Medicine  \textbf{104} (2020), \url{www.scopus.com}, cited By :125

\bibitem{weng2019}
Weng, Z.: From conventional machine learning to automl. In: Journal of Physics:
  Conference Series. vol.~1207 (2019), \url{www.scopus.com}, cited By :9

\bibitem{zeiler2012adadelta}
Zeiler, M.D.: Adadelta: an adaptive learning rate method. arXiv preprint
  arXiv:1212.5701  (2012)

\bibitem{zoph2018learning}
Zoph, B., Vasudevan, V., Shlens, J., Le, Q.V.: Learning transferable
  architectures for scalable image recognition. In: Proceedings of the IEEE
  conference on computer vision and pattern recognition. pp. 8697--8710 (2018)

\bibitem{zoller2021}
Zöller, M.., Huber, M.F.: Benchmark and survey of automated machine learning
  frameworks. Journal of Artificial Intelligence Research  \textbf{70},
  409--472 (2021), \url{www.scopus.com}, cited By :30

\end{thebibliography}

\end{document}